\documentclass[10pt]{article}

\usepackage[margin=1in]{geometry}
\usepackage{microtype}
\usepackage{times}
\usepackage{graphicx}
\usepackage{subcaption}
\usepackage{booktabs} 
\usepackage[round,authoryear]{natbib}
\usepackage{amsmath}
\usepackage{amssymb}
\usepackage{mathtools}
\usepackage{amsthm}
\usepackage[most]{tcolorbox}
\usepackage{listings}
\usepackage{xcolor}
\usepackage{float}
\usepackage{multirow}
\usepackage{hyperref}
\usepackage[capitalize,noabbrev]{cleveref}
\definecolor{linkblue}{RGB}{0,0,130}
\hypersetup{
  colorlinks=true,
  linkcolor=linkblue,
  citecolor=linkblue,
  urlcolor=linkblue,
  pdftitle={Counsel: A Meta-Evaluation Dataset for Agentic Tasks},
  pdfauthor={Sashank Pisupati, Henry Broomfield, Eujeong Choi, Antonia Calvi, Charlie Wang, Roman Engeler, Max Bartolo, Patrick Lewis}
}
\newcommand{\hficon}{\raisebox{-0.18em}{\includegraphics[height=1.1em]{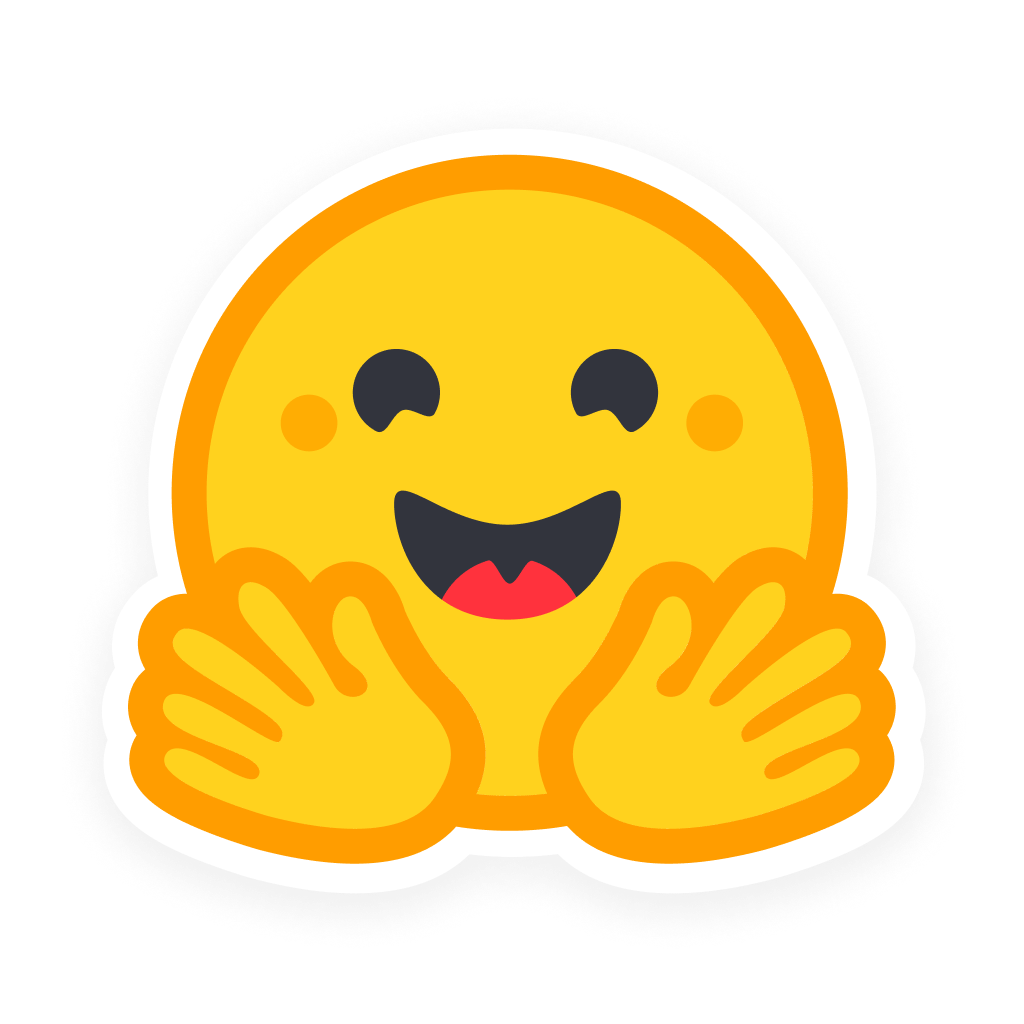}}}

\lstdefinestyle{prompt}{
  basicstyle=\ttfamily\small,
  breaklines=true,
  breakatwhitespace=true,
  columns=fullflexible,
  keepspaces=true,
  showstringspaces=false,
  frame=single,
  framerule=0.5pt,
  rulecolor=\color{black},
  xleftmargin=0.6em,
  xrightmargin=0.6em,
  framexleftmargin=0.6em,
  framexrightmargin=0.6em,
  aboveskip=0pt,
  belowskip=0pt,
  breakautoindent=false,
  breakindent=0pt
}

\theoremstyle{plain}

\theoremstyle{definition}

\theoremstyle{remark}

\begin{document}

\twocolumn[
\begin{center}
{\LARGE\bfseries Counsel: A Meta-Evaluation Dataset for Agentic Tasks\par}
\vspace{1.2em}
{\large
Sashank Pisupati$^{1,*}$ \quad Henry Broomfield$^{1,*}$ \quad Eujeong Choi$^{2,*}$ \quad Antonia Calvi$^{3,*}$ \\
Charlie Wang$^2$ \quad Roman Engeler$^1$ \quad Max Bartolo$^4$ \quad Patrick Lewis$^2$ \\
}
\small
$^1$Atla AI \quad $^2$Cohere AI \quad $^3$Mistral AI \quad $^4$Google DeepMind \\
London, UK \\
$^*$Equal contribution. Correspondence: \href{mailto:sashank.pisupati@gmail.com}{sashank.pisupati@gmail.com}, \href{mailto:henry.a.broomfield@gmail.com}{henry.a.broomfield@gmail.com}
\end{center}
\vspace{1.5em}

\begin{abstract}
  As agentic systems tackle increasingly complex multi-step tasks, evaluating their trajectories presents a major bottleneck---human annotation of a single trajectory on popular agentic benchmarks can take hours, making it difficult to scale evaluations for measuring performance or curating training data. This has driven widespread reliance on automated approaches such as LLM-as-a-judge (LLMJ) to critique agents at the process and outcome-levels at scale, however, the soundness of LLMJ critiques often goes unmeasured. Here, we introduce Counsel, the first public dataset of meta-evaluations for agentic tasks. Counsel consists of process-level critiques from open-weight LLMJs on two agent benchmarks: $\tau$-bench (customer support agents) and DA-Code (coding agents), and human meta-evaluations of these critiques. Human annotators label critiques on each flagged error as ``spot on", ``correct location but poor reasoning", or ``should not have flagged", achieving reliable inter-annotator agreement (Krippendorff`s $\alpha$ of 0.78). The resulting dataset stratifies LLMJ critiques by human alignment across both error location within a trajectory and reasoning quality, serving as valuable data to calibrate, improve, or train LLMJs for agents. Comparing open-weight judges, we find that more capable judge models and more reasoning effort both enabled improved human agreement, with the strongest judge reaching $\sim$ 88 \% agreement on location and $\sim$ 65 \% on reasoning. Counsel is generated using open-weight models and is permissively licensed for broad community use, which we hope will enable rigorous study and improved alignment of LLM-based evaluators for agentic systems.
\end{abstract}

\begin{center}
\small \hficon\ \href{https://huggingface.co/datasets/AtlaAI/counsel}{\textbf{Dataset:} \texttt{huggingface.co/datasets/AtlaAI/counsel}}
\end{center}
\vspace{1em}
]

\section{Introduction}

Agentic systems are being applied to ever longer and more complex tasks; by one estimate, the 50\% task completion time horizon is doubling every 7 months \citep{kwa2025measuring}. This increase in complexity brings with it an equivalent increase in the human cognitive burden required to annotate agent trajectories, a critical step in comprehensively evaluating performance or curating new training data. For instance, the popular SWE-bench \citep{jimenez2023swe} contains tasks that an experienced software engineer is expected to take $\sim$1 hour to complete, and agent trajectories on this benchmark take human annotators $\sim$2 hours to annotate \citep{deshpande2025trail}. While many agent benchmarks utilize programmatic or verifiable criteria to evaluate trajectory outcomes, such criteria are much rarer at the step level, and the severe annotator burden makes human feedback on fuzzy, qualitative or non-verifiable dimensions extremely challenging to scale. 

This has led many practitioners to adopt automated approaches to evaluating agents such as LLM-as-a-judge (LLMJ), which make use of LLMs to generate natural language critiques of agent trajectories and emulate human judgments at both the process-level (i.e. judging every step of an agent's trajectory) and the outcome-level (i.e. judging the final result of an agent's trajectory). LLMJs are used across the board, from data curation and training \citep{guo2025deepseek, team2025kimi} to deployment of agentic models. 

In particular, LLMJ critiques are increasingly used as diagnostic artifacts, to gain qualitative insight into agent performance \citep{grace2026demystifyingEvalsAgents, ou2025agentdiagnose} since they provide much richer descriptions of agent behavior than numeric scores. However, despite their widespread adoption, the soundness of LLMJ critiques of agent trajectories often goes unmeasured, with most attempts at \textit{meta-evaluation}---i.e. evaluating the outputs of an evaluator---restricting themselves to comparing final judgments alone with humans \citep{deshpande2025trail}. This lack of meta-evaluation data means that attempts at measuring or improving the quality of LLMJs can only indirectly target critiques---hoping that improving judgment accuracy leads to higher critique quality. 

\begin{figure*}[ht!]
  \centering
  \makebox[\textwidth][c]{
    \includegraphics[width=0.9\textwidth]{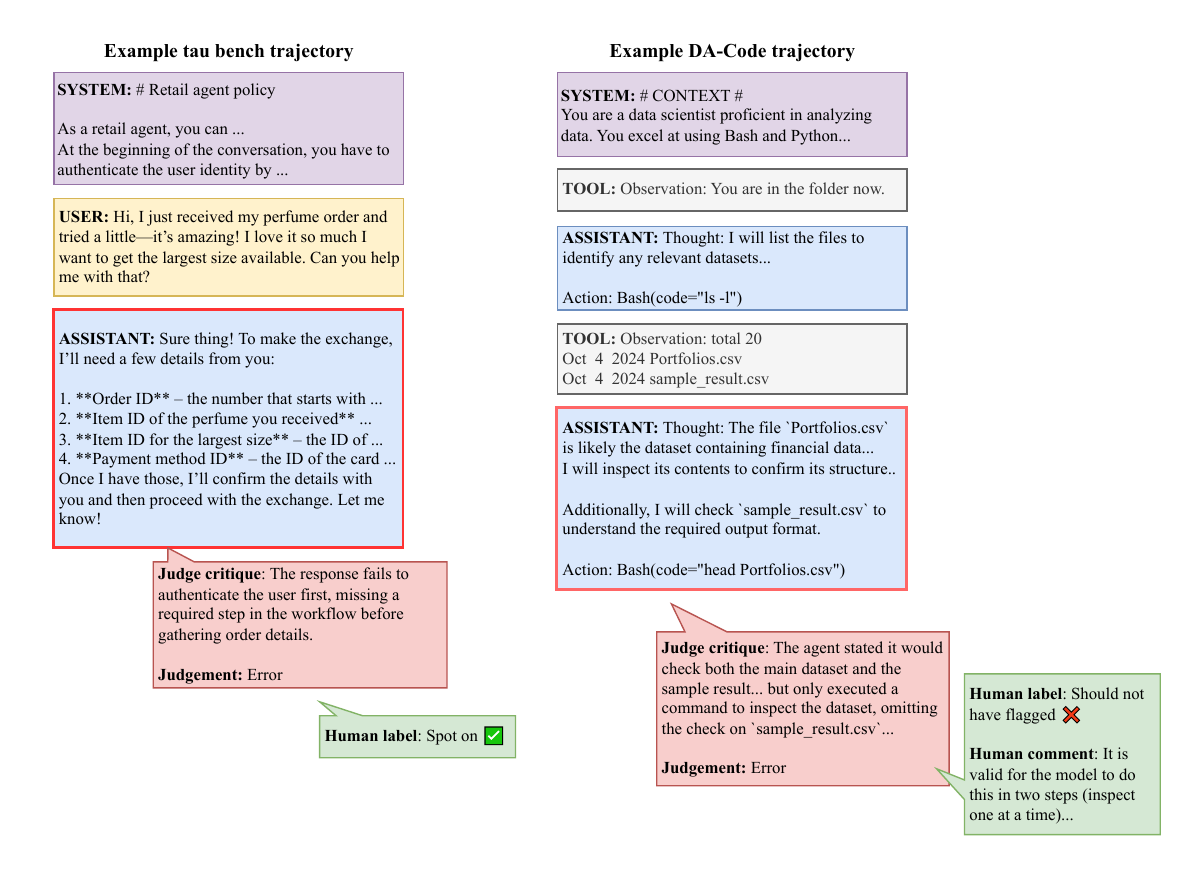}
  }
  \caption{\textbf{Example agent trajectories, judge critiques, and human meta-annotations from Counsel.} Each side represents a trajectory with multiple spans, where each box shows the output generation of a span that is conditioned on information in only preceding boxes. \textbf{Left}: A trajectory from $\tau$-bench, a customer support benchmark. In this interaction an LLMJ (red) correctly flags the error location and reasons correctly about what the error is, hence, it is labeled as "Spot on" by the meta-annotator (green). \textbf{Right}: A trajectory from DA-Code, a coding benchmark. In this interaction, an LLMJ incorrectly and prematurely flags a reasonable intermediate step by an agent, and is labeled as "Should not have flagged" by the meta-annotator.}
  \label{fig:examples}
\end{figure*}

In this work, we present Counsel, the first public dataset of meta-evaluations for agentic tasks. This dataset contains high quality human meta-evaluations of process-level LLMJ critiques on two agent benchmarks: $\tau$-bench (customer support) and DA-Code (coding). We use open-weight LLMJs to critique an open-weight agent's trajectory and flag whether each step contains an error and why. We then ask human annotators to label flagged steps on both error location identification correctness, and correctness of critique explanation. We find that annotators achieve high inter-annotator agreement (Krippendorff's alpha 0.78), and that more capable judge models and higher reasoning effort both enable higher human agreement, with the strongest judge achieving 88\% agreement on location and 65\% on reasoning.

We intend Counsel to serve as a resource for studying and improving LLMJs. In particular, the dataset enables three primary uses:

\textbf{(1) Evaluating critique quality of judges.}
Human meta-annotations provide a reference for assessing whether judges correctly localize errors and whether the critiques contain high quality reasoning, enabling evaluation of LLMJs beyond agreement on final scores.

\textbf{(2) Training and improving meta-judges.}
The dataset can be used to optimize meta-judges that score, rank, or filter judge outputs, supporting the selection of higher-quality critiques.

\textbf{(3) Training and improving judges.}
Meta-annotations can serve as supervision or reward signals for directly optimizing judges to produce higher quality critiques.

\section{Related work}
Our primary contribution is a public and permissive dataset containing high quality human meta-evaluations of LLMJ \textit{critiques} as well as \textit{judgments} across two agentic domains. In addition to its immediate utility in benchmarking LLMJ quality on agentic tasks, our dataset can be used to train meta-judges which can then be used to improve the quality of LLMJs for agents. 

Prior work on benchmarking LLMJ performance has largely been restricted to meta-evaluating final judgments (scores/preferences) by comparing them to human judgments (e.g. JudgeBench \citep{tan2024judgebench}, RewardBench \citep{malik2025rewardbench}), and this continues to be the case in agentic domains in works such as AgentRewardBench \citep{lu2025agentrewardbench}, Agent-as-a-Judge \citep{zhuge2024agent} and MAST \citep{cemri2025multi}. These, alongside a number of ``golden" datasets containing human judgments of agent trajectories, such as TRAIL \citep{deshpande2025trail}, offer valuable training and evaluation data for aligning LLMs' final judgments to those of humans, but do not offer meta-evaluations of critiques. Hence, while many approaches to training LLMJs use reasoning and critiques to improve performance \citep{wang2025direct, alexandru2025atla, whitehouse2025j1, chen2025rm, guo2025reward}, these still rely solely on the correctness of the final judgment as a measure of critique quality, and hence a training signal. Recent attempts at improving the reasoning capabilities of LLMs have shown that training a verifier model using human data---then using it to score reasoning chains from the original model---is much more data efficient than training directly on human data \citep{liu2023tinygsm}. In a similar spirit, the human meta-evaluations in our dataset offer valuable training data for improving the critiquing ability of LLMJs, by training human-aligned \textit{meta-judges}. Such ``meta-judge" approaches offer richer feedback signals than relying on the correctness of final judgments alone, and have been shown to improve LLMJs even with off-the-shelf meta-judges \citep{wu2025meta, li2025leveraging}.

\section{Methods}

We begin by defining core terminology, then describe the procedures used to generate agent trajectories, judge critiques, and human meta-annotations.

\subsection{Terminology}

In this work, we refer to \textbf{trajectories}, \textbf{spans}, \textbf{judgments}, and \textbf{critiques} that we define as follows:

\begin{itemize}
    \item \textbf{Trajectories} are the ordered steps that define the cohesive context that the agent and the environment generate during execution of a task. Concretely, it is the combined list of messages in an OpenAI-compatible chat completions format, including user utterances, assistant utterances and tool calls, and tool outputs. 
    \item \textbf{Spans} correspond to single steps of the model within a trajectory. Specifically, a span is a model or tool invocation that receives the prior context of the trajectory as the request’s input, and the output of the request, which is the result of a tool call or the message of the model’s generation (content to the user or JSON to call a tool). See \Cref{fig:examples} for examples.
    \item \textbf{Judgments} are categorical predictions from an LLMJ. \Cref{sec:generating-judge-evaluations} outlines the binary indicator of error presence used for judgments within Counsel.
    \item \textbf{Critiques} are textual feedback from an LLMJ that provide a human-readable explanation about why a judgment was given. See \Cref{fig:examples} for examples. The \textbf{judge output} is a generation of the critique and the judgment in that order.
\end{itemize}

\subsection{Agent environments}

Two popular agentic environments are selected, namely $\tau$-bench \citep{Yao2024} and DA-Code \citep{Huang2024}. These two datasets encompass two of the most common agentic use-cases: customer service, and code generation. Here, we describe some of the specific configurations applied to data collection for each of these environments.

\subsubsection{$\tau$-bench}

The Tool-Agent-User (TAU or $\tau$) interaction benchmark, or $\tau$-bench, evaluates agents on multi-turn tool-agent-user interactions in real-world domains. It simulates dynamic conversations between a user and an agent that must use domain-specific API tools and adhere to business rules and guidelines (e.g. in retail or airline customer support). At the end of a conversation, the benchmark grades the agent’s performance by comparing the final state of the underlying database to a known goal state, alongside whether the agent communicated required confirmations (e.g. a price that a user must pay or confirmation of successful interaction). This outcome-centric evaluation allows for objective, programmatic measurement of whether the user’s request was fulfilled, providing an average reward over a set of agent attempts.

Specifically, we focus solely on the \textit{test} examples of the \textit{retail} subset as this is the most widely accepted, largest subset of $\tau$-bench, with human annotations of ground-truth desired database state. This totals 115 tasks that could be attempted by agents.

\subsubsection{DA-Code}

The Agent Data Science Code Generation benchmark, or DA-Code, is designed to evaluate agents on realistic, end-to-end data-science workflows. It consists of complex tasks spanning data cleaning, exploratory analysis, modeling, and multi-file code execution across Python, SQL, and Bash within a fully instrumented sandboxed environment, enabling fine-grained assessment of an agent’s planning, reasoning, and coding abilities. An agent’s task attempt is graded by executing the agent’s code within the sandbox and comparing the resulting tables, visualizations, and model outputs against ground-truth expectations, enabling objective, programmatic evaluation of task completion.

For our experiments, we restrict evaluation to the \textit{Data Insights} (DI) and \textit{Data Manipulation} (DM) subsets. These are chosen through manual inspection as the most reliable and internally consistent portions of the benchmark for generating quality agent trajectories and judgments. See \Cref{sec:da-code-underspecification} for further quality assurance details. This results in 50 DA-Code tasks that could be attempted.

\subsection{Trajectory generation}

Having established the agentic environments in which we wish to study LLMJ performance, we generate agent trajectories by rolling out tasks with an agent model. This subsection details the choice of agent models and rollout configuration. The following subsection describes the choice of LLMJ models and the procedure used to judge agent steps.

We select agent, judge, and user models that are sufficiently capable to produce realistic failure modes and to generate critiques that serve as strong reference signals for human meta-evaluation and downstream training. Additionally, to permit open and permissive community use of the dataset, we restrict both agents and judges to open-weight models.

\subsubsection{Agent models}

Agents are chosen to induce heterogeneous failure modes---spanning different model families, scales, and reasoning configurations---so that judge critiques are not dominated by a single policy class. We also vary the reasoning capability to avoid failure regimes dominated by either under-thinking (leading to superficial errors) or unrealistic configurations utilizing excessive deliberation and latency.

\begin{itemize}
    \item \textbf{GPT-OSS-20B} (medium reasoning): A recently released 20B-parameter mixture of experts (MoE), 3.6B active parameters, open-weight, reasoning model designed for agentic workflows \citep{OpenAI2025}. It is available under an Apache-2.0 license with competitive performance on agent benchmarks, including $\tau$-bench retail. Medium reasoning level is chosen to balance realistic agent latency with sufficient deliberation to produce non-trivial, heterogeneous failure modes for meaningful judge evaluation.
    \item \textbf{Qwen3-235B-A22B-Instruct-2507} (no reasoning): A recent 235B-parameter MoE model with 22B active parameters released as an open-weight frontier non-reasoning model under an Apache-2.0 license \citep{Yang2025}. Its broad applicability makes it appropriate for fine-tuning, and system 1 thinking permits lower latency for agent-environment interactions despite the model’s larger overall size.
\end{itemize}

\subsubsection{Generating agent trajectories}\label{sec:generating-agent-trajectories}

Minimal changes are made to the respective $\tau$-bench and DA-Code repositories to enable querying of models through Together AI’s inference endpoint, and default settings are used where possible. In $\tau$-bench specifically, we invoke the standard \texttt{ToolCallingAgent} and pair it with the capable simulated user model, Qwen3-235B-A22B-Instruct-2507, so that observed failures and resulting judge critiques are attributable to the agent rather than confounded by user-side errors or underspecification, while continuing the open-license theme.

\subsection{Judgment Generation}

\subsubsection{Judge models}

Our goal in selecting judge models is to induce systematic variation in judgment behavior along two central axes: model family and reasoning effort. This variation increases the diversity of judgments in the dataset, reducing dependence on any single model’s failure modes while ensuring coverage of a broad range of error localizations and explanations.

\begin{itemize}
    \item \textbf{GPT-OSS-120B}: A 120B-parameter mixture-of-experts model with 5.1B active parameters, released as an open-weight model under the Apache-2.0 license \citep{OpenAI2025}. We generate judge outputs in both \textit{low} and \textit{high} reasoning settings to capture a range of inference behaviors.
    \item \textbf{Qwen3-235B-A22B-Instruct-2507} (no reasoning): A widely used, highly capable open-weight frontier model with strong instruction-following and code understanding, making it a competitive non-reasoning baseline for judging agent behavior.
\end{itemize}

Employing both same-family (GPT$\leftrightarrow$GPT, Qwen$\leftrightarrow$Qwen) and cross-family judge–agent pairings (GPT$\leftrightarrow$Qwen) enables analysis of self-preference bias \citep{Wataoka2024}, a well-documented phenomenon in prior work where models are more lenient toward their own generations.

\subsubsection{Generating judge evaluations}\label{sec:generating-judge-evaluations}

Each model call span in a trajectory is evaluated by an LLM-as-a-Judge. The LLMJ receives the same trajectory information as the agent that took the step (including the tools that the agent had access to), and the agent’s output at that step. The judge does not have privileged access to future information from subsequent spans, and each span evaluation is independent (i.e., it did not receive previous spans’ judge outputs). This mirrors the online setting in which critiques must be produced from available context and retains the problem structure needed for training effective guardrail judges and process reward models.

The LLMJ’s prompt and evaluation criteria are detailed in \cref{fig:judge-prompt}. It is tuned to open-code errors that could lead to failure of its task. While a static error type taxonomy is provided, this taxonomy is only to help the evaluator understand its task and think through potential failure modes that it could ascribe. The judge is asked to provide some critique (on top of its internal reasoning, if the judge model is capable of reasoning), while the critique is requested to be concise, human-digestible, and specific to the particular open-ended failure mode that the agent may have exhibited in its step. Lastly, a judgment is received that is a binary indicator of error presence. All judge queries are made through Together AI’s inference endpoints that enable structured outputs for greater reliability of schema conformance.

\subsection{Human annotation}

We collect human meta-judgments to evaluate the quality of automated judge model outputs on agentic benchmark trajectories. Specifically, annotators assess both the correctness of error location identification and the accuracy of critiques generated by LLMJ models performing step-wise evaluation. There is also an optional text field to provide an open-ended comment. Although not a focal point or quality controlled, we release these comments alongside the dataset.

\subsubsection{Annotation Scope}

Human annotators evaluated the judge outputs produced by the three LLMJs on agentic trajectories from both $\tau$-bench and DA-Code. Annotators only review complete trajectories that contained at least one error flagged by the judge models. This is to reduce annotation burden and enable scaling up annotator throughput, as it only requires annotators to review potential issues flagged by judges --- whereas reviewing unflagged trajectories would require annotators to also perform the original, more cumbersome evaluation task. This means that Counsel's design of meta-annotated critiques \textit{prioritizes precision of judges’ evaluations over recall}.

Human annotators provide feedback on each flagged span using privileged information of the full trajectory, including past and future judge span outputs, unlike judges that only had access to past trajectory information. This approach ensures annotators have sufficient context to accurately assess whether errors are correctly identified and whether the judge's critique is sound.

\subsubsection{Annotation Schema}

We develop a unified three-way labeling scheme that jointly evaluates both the location and critique soundness of each judge output:

\begin{itemize}
    \item \textbf{Spot On}: Both the error location and the critique are correct.
    \item \textbf{Poor Reasoning but Correct location}: The judge correctly identifies where an error occurred, but provides an incorrect or inadequate critique for why it is an error.
    \item \textbf{Should Not Have Flagged}: The judge incorrectly flags this location as containing an error (both location and critique are wrong).
\end{itemize}

This schema is designed after considering that the combination of poor location with good reasoning is not a meaningful category in practice. If the error location is incorrect, the associated reasoning is not considered correct in context, since it explains a non-existent error.

\subsubsection{Annotator Selection and Training}

Three skilled professional data science annotators, each with over 10 years of experience in data science and natural language processing, serve as annotators for this task. Prior to annotation, annotators undergo training that includes: (1) review of detailed annotation guidelines covering the three-way labeling schema and edge case handling, (2) familiarization with the annotation platform interface and workflow, and (3) practice annotations on sample trajectories followed by group discussion to calibrate understanding and resolve ambiguities.

\subsubsection{Annotation process}

Annotations are conducted on an interface that displays full agent trajectories along with judge model output on the left side, and the annotation interface on the right side, allowing annotators to review flagged errors in their complete context while providing their meta-judgments including an optional meta-critique.

Trajectories are randomly assigned to annotators to ensure balanced distribution across judge models and benchmarks. To establish inter-annotator agreement, an initial batch of 15 trajectories receives triple annotation (all three annotators independently labeled the same trajectories). Following this calibration phase, the remaining trajectories are divided among annotators for single annotation.

Quality control is maintained throughout the annotation process through periodic reviews conducted by a senior annotator, who examines annotation decisions and provides feedback to ensure consistency with the annotation guidelines and alignment across annotators.

Average annotation time varies by benchmark complexity, with approximately 20 minutes per trajectory for $\tau$-bench and 30 minutes per trajectory for DA Code, reflecting the additional complexity of evaluating code-based agent interactions.

\subsubsection{Inter-Annotator Agreement}

To assess annotation quality, we calculate inter-annotator agreement with the triple-annotated sample batches using Krippendorff's alpha \citep{krippendorff2004reliability}. The annotators achieve substantial agreement with $\alpha\approx0.78$, indicating high reliability in the meta-judgment task despite its inherent complexity.

\section{Dataset analysis}

In this section, we present descriptive statistics of Counsel, characterizing agent trajectories, judgments, and meta-annotations. These analyses highlight key properties of the dataset arising from our design choices and inform its use for studying and improving LLMJs.

Some examples of critiques, and human meta-annotations of the critique quality are provided in \Cref{fig:examples}.

\begin{table}[t]
  \caption{\textbf{Number of unique trajectories} in Counsel, i.e.\ those where any judge marked an error on any span. Recall that GPT-OSS-20B is not used for DA-Code. See \cref{tab:human-meta-annotated-judgment-quality} for the number of annotations.}
  \label{agent-performance-table}
  \begin{center}
    \begin{small}
      \begin{sc}
        \begin{tabular}{lccc}
          \toprule
          Agent model
          & GPT-OSS-20B
          & \begin{tabular}[c]{@{}c@{}}
              Qwen3
            \end{tabular}
          & Total \\
          \midrule
          da-code          & -- & 40 & 40  \\
          $\tau$-bench retail & 90 & 95 & 185 \\
          \midrule
          Total            & 90 & 135 & 225 \\
          \bottomrule
        \end{tabular}
      \end{sc}
    \end{small}
  \end{center}
  \vskip -0.1in
  \label{tab:traces}
\end{table}

\subsection{Agent trajectories}

\cref{fig:agent-steps-and-tokens} details the number of steps taken by the agent. DA-Code typically results in fewer agent steps per trajectory than $\tau$-bench, but has a greater number of output (not including reasoning) tokens per agent step, due to greater task complexity.

For $\tau$-bench, Kolmogorov-Smirnov tests of distributional mismatch are performed between characteristics of Qwen3 and GPT-OSS-20B as agents. This is particularly important for assessing the extent of biases, particularly the length bias of LLMJs \citep{Park2024}, for subsequent analyses.

A test comparing the number of spans per trajectory yields no significant difference ($p=0.929$). Comparing the number of output (not including reasoning) tokens per step identifies greater generation length by Qwen3 ($p=0.0001$); however, the effect size of the difference in mean length (Cohen’s d = 0.329) is small, and likely would not have a great impact. As such, \textit{length bias is not a concern}.

\subsection{Judgments}

\begin{figure*}[t]
  \centering
  \makebox[\textwidth][c]{
    \includegraphics[width=0.9\textwidth]{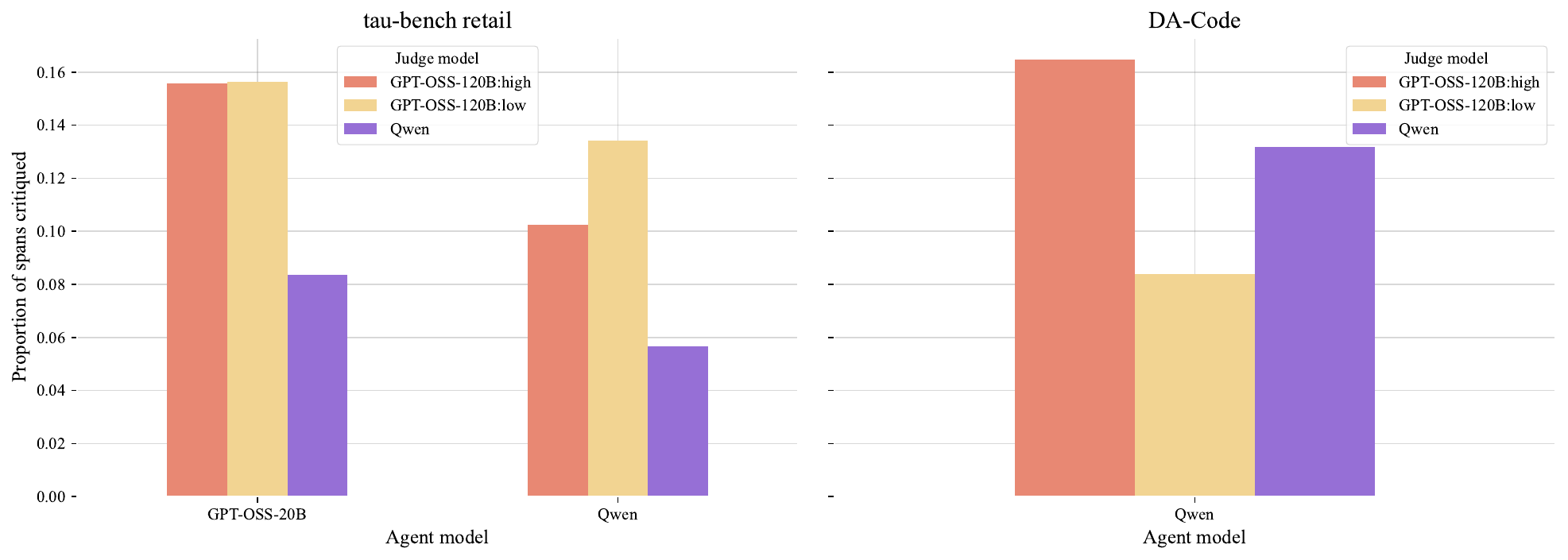}
  }
  \caption{\textbf{Judge critique rates} across agents, benchmarks, and judge models. The figure shows the proportion of agent spans flagged as containing an error by each judge model, stratified by agent model for $\tau$-bench retail (\textbf{left}) and DA-Code (\textbf{right}).}
  \label{fig:judge-critique-rate}
\end{figure*}

We examine the propensity of judge models to evaluate the agent models in \cref{fig:judge-critique-rate}. According to the judges, Qwen3 is a more performant agent, making fewer errors than GPT-OSS-20B on $\tau$-bench. Here, GPT-OSS-120B as a judge is more strict than Qwen3 as a judge.

To identify self-preference bias, the judge outputs are categorized into those that were of the same model family or a different model family. As an example, GPT-OSS-120B evaluating GPT-OSS-20B would be considered the same model family, while GPT-OSS-120B evaluating Qwen3 would be a different model family critique. \cref{fig:self-bias} breaks down these results, with one-sided normal tests for proportions identifying \textit{no significant self-preference bias of any of the judges}.

\subsection{Human meta-annotations}

\begin{figure*}[t]
  \centering
  \makebox[\textwidth][c]{
    \includegraphics[width=0.9\textwidth]{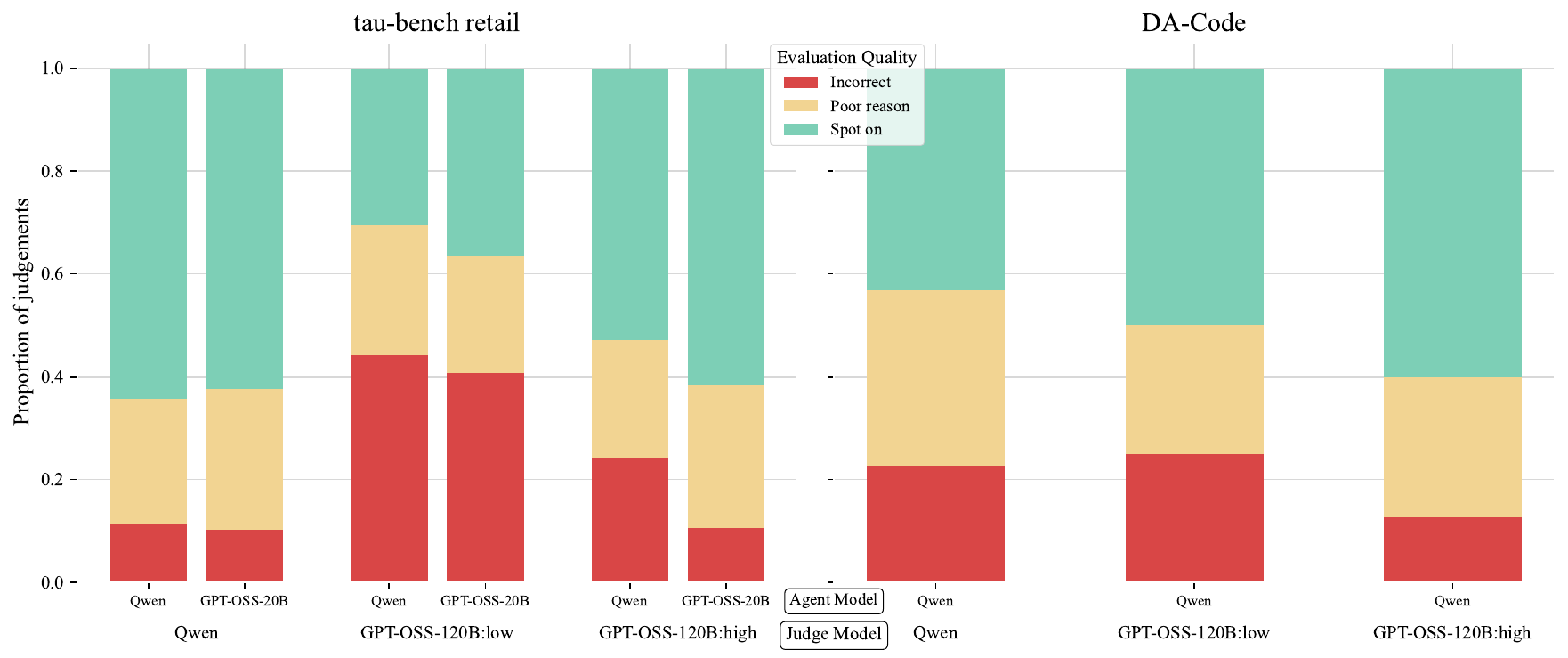}
  }
  \caption{\textbf{Human meta-annotated quality of judge outputs}. Proportion of critique and judgment labeled by human annotators as Spot On, Poor Reasoning (correct location), or Should Not Have Flagged, broken down by agent model, and judge model for each of $\tau$-bench retail (\textbf{left}), and DA-Code (\textbf{right}).}
  \label{fig:judge-precision}
\end{figure*}

We evaluate the accuracy of judge outputs using the incorrect, poor reasoning, and spot on human meta-annotations. The overall number of judge outputs and their human meta-annotated quality from each agent-judge-benchmark combination are detailed in \cref{tab:human-meta-annotated-judgment-quality}, and visually depicted in \cref{fig:judge-precision}. Here, we outline some key findings.

\begin{table*}[t]
  \caption{\textbf{Human meta-annotated judge output quality} across benchmark environments, agent models, and judge models. Counts correspond to the number of critiques labeled as \textit{Spot On}, \textit{Poor Reasoning (correct location)}, or \textit{Should Not Have Flagged}.}
  \label{tab:human-meta-annotated-judgment-quality}
  \begin{center}
    \begin{small}
      \begin{sc}
        \begin{tabular}{llcccc}
          \toprule
          \multicolumn{2}{l}{\textbf{Benchmark}} 
            & \textsc{DA-Code} 
            & \textsc{$\tau$-bench retail} 
            & \textsc{$\tau$-bench retail}
            & \textbf{Total} \\
          \multicolumn{2}{l}{\textbf{Agent model}} 
            & Qwen 
            & GPT-OSS-20B 
            & Qwen
            &  \\
          \midrule
          \textbf{Judgment} & \textbf{Judge model} & & & & \\
          \midrule
          \multirow{3}{*}{Spot On}
            & GPT-OSS-120B (high) & 33 & 134 & 83 & 250 \\
            & GPT-OSS-120B (low)  & 14 & 80  & 63 & 157 \\
            & Qwen3               & 19 & 73  & 56 & 148 \\
          \cmidrule(lr){2-6}
            & \textit{Total}      & 66 & 287 & 202 & 555 \\
          \midrule
          \multirow{3}{*}{Poor Reasoning}
            & GPT-OSS-120B (high) & 15 & 61 & 36 & 112 \\
            & GPT-OSS-120B (low)  & 7  & 50 & 52 & 109 \\
            & Qwen3               & 15 & 32 & 21 & 68  \\
          \cmidrule(lr){2-6}
            & \textit{Total}      & 37 & 143 & 109 & 289 \\
          \midrule
          \multirow{3}{*}{Should Not Have Flagged}
            & GPT-OSS-120B (high) & 7  & 23 & 38 & 68  \\
            & GPT-OSS-120B (low)  & 7  & 89 & 91 & 187 \\
            & Qwen3               & 10 & 12 & 10 & 32  \\
          \cmidrule(lr){2-6}
            & \textit{Total}      & 24 & 124 & 139 & 287 \\
          \midrule
          \textbf{Total} &  & 127 & 554 & 450 & 1131 \\
          \bottomrule
        \end{tabular}
      \end{sc}
    \end{small}
  \end{center}
  \vskip -0.1in
\end{table*}

\textbf{Judge accuracy varies with agentic domain.} Qwen3’s critique precision (proportion of judgments that are “Spot On”) is 43\% on DA-Code and 63\% on $\tau$-bench retail (micro-average of agent models). On the other hand, GPT-OSS-120B:high achieves 60\% critique precision on DA-Code and 58\% on $\tau$-bench retail, together highlighting that different models' critique precision varies according to the specific challenges that an environment poses.

To reiterate, this report mainly analyzes the precision of judge outputs; however, we may make claims about relative recall by comparing true positives across judges, since the total number of true labels remains the same (\Cref{tab:human-meta-annotated-judgment-quality}). For example, on DA-Code, GPT-OSS-120B:high has a $33/19\rightarrow74\%$ greater recall than Qwen3.

\textbf{Greater reasoning effort improves critique quality.} GPT-OSS-120B:high generates fewer evaluations marked ``poor reasoning" or ``should not have flagged" than its low reasoning counterpart across both benchmarks, suggesting that critique quality does benefit from increased reasoning compute.

\section{Discussion}

Counsel is the \textit{first meta-evaluation dataset for agentic tasks}, introducing labels on LLM-as-a-Judge output quality across coding and customer service environments. By pairing real agent trajectories with step-level LLMJ critiques with human meta-judgments (“spot on”, “poor reasoning in the correct location”, “should not have flagged”), the dataset supports models that are not only accurate, but also usefully diagnostic: they identify the right failure point and articulate critiques that can drive debugging, guardrails, or learning. In addition to the dataset, the proposed approach provides a practical and scalable foundation for improving evaluation in agentic systems, which are an increasingly prominent application setting and remain challenging to observe directly at scale.

\subsection{Limitations and future work}

The focus of Counsel is the quality of judges' critiques where they \textit{do identify} errors, but not where the judge \textit{fails to raise} an issue. Hence, the dataset supports the measurement and improvement of the precision of judges and the quality of their critiques where something was flagged, but not the recall of error localization. Only two agentic environments and model families were tested, which limits how much the dataset may generalize across other dissimilar agentic domains.

While Counsel provides a resource to calibrate, improve, and train LLMJ systems, this work primarily focuses on constructing and characterizing the dataset and outlining its intended uses. A key direction for future work is to systematically study downstream applications, including how meta-judgments can be used as learning signals to improve judge behavior.

In \cref{sec:experiment-1}, we present a preliminary experiment illustrating one such use, where meta-annotations are leveraged as few-shot examples to shape judge outputs. More broadly, Counsel enables the development of \textit{automated meta-judges} that score, rank, or filter judge critiques---selecting higher-quality evaluations, filtering spurious flags, and calibrating strictness---which we leave for future investigation.

\section*{Impact statement}

By enabling higher quality diagnostic critiques of agent behavior, this work can support safer deployment, better debugging, and more effective training of agents in complex, real-world tasks. More reliable evaluators may reduce reliance on costly human oversight while improving transparency and accountability in agent decision-making. As with any evaluation technology, there is a risk that automated judges could be over-trusted or misapplied outside their validated domains; we therefore emphasize that Counsel is intended to complement, not replace, human judgment. We release the dataset using open-weight models and permissive licensing to encourage broad, responsible research into scalable and aligned evaluation methods.

\bibliography{bibliography}
\bibliographystyle{plainnat}

\newpage
\appendix
\onecolumn

\newpage
\section{Additional DA-Code filtering}\label{sec:da-code-underspecification}

An underspecification problem with some of the tasks (see \cref{sec:da-code-underspecification}) led to an additional quality-assurance process. Two frontier models (OpenAI’s o3 and Anthropic’s Claude-3.5-Sonnet) were each run as agents through all \textit{Data Insights} and \textit{Data Manipulation} tasks. Any task where fewer than 50\% of its samples were completed correctly are discarded.

Although this biases tasks to an easier subset, it is deemed acceptable for a couple of reasons. Firstly, it results in fewer false negative agent runs from the benchmark’s internal evaluation criteria that could mislead a human meta-annotator and reduce meta-annotation quality. Secondly, the focus of the dataset is to measure evaluation quality of the judges, thus the simpler subset augments the number of agent attempts that could lead to false positive error flags by the judges.

\begin{figure}[H]
\centering
\begin{tcolorbox}[title=Task description]
\small
Calculate a set of summary statistics on the purchase data, broken down by \texttt{device} (Android or iOS) and \texttt{gender} (Male or Female). These summary statistics should include the total purchases, average purchase amount, and purchase count for each group. Fill the corresponding blanks in \texttt{purchase\_summary.csv} with these summary statistics, ensuring the format matches the provided template.
\end{tcolorbox}

\vspace{0.5em}

\begin{tcolorbox}[title=Agent output]
\small\ttfamily
device,gender,mean price,median price,std price \\
and,F,400.75,299.0,179.98 \\
and,M,416.24,499.0,195.0 \\
iOS,F,404.44,299.0,181.52 \\
iOS,M,405.27,299.0,196.84
\end{tcolorbox}

\vspace{0.5em}

\begin{tcolorbox}[title=Gold output]
\small\ttfamily
device,gender,mean price,median price,std price \\
and,F,400.7475035663338,299.0,179.98437839269033 \\
and,M,416.2373081463991,499.0,195.00152018707635 \\
iOS,F,404.4353296426774,299.0,181.52495249573712 \\
iOS,M,405.27240143369175,299.0,196.8431973487182
\end{tcolorbox}

\caption{
\textbf{Example of task underspecification in DA-Code}. The task (\textbf{top}) does not specify numerical precision or rounding requirements. The agent produces correctly computed prices rounded to two decimal places (\textbf{middle}), while the benchmark gold output (\textbf{bottom}) retains full floating-point precision. Despite the agent’s output being semantically correct and consistent with common data reporting practices, the mismatch in formatting leads to an incorrect evaluation under the benchmark’s matching criteria. This type of underspecification motivated additional quality control when selecting DA-Code tasks for meta-evaluation.
}
\label{fig:da-code-underspecification}
\end{figure}

\section{Judge prompt}

\begin{figure}
    \centering
    \setlength{\abovecaptionskip}{2pt}
    \setlength{\belowcaptionskip}{0pt}
    \begin{minipage}{0.98\linewidth}
    \lstset{
      style=prompt,
      basicstyle=\ttfamily\scriptsize
    }
\begin{lstlisting}
You are a QA analysis expert tasked with evaluating whether an AI agent suffers from an error in its current step. Think about these types of errors to help guide your analysis:

Example agentic error types:
- user_interaction_error: the agent returned incorrect information to the user based on its tool outputs or compared to its instructions.
- reasoning_error: the agent has generated incorrect or incomplete logic, flawed reasoning, incorrect code, hallucinated information that didn't exist, or did not properly follow its instructions while reasoning.
- tool_use_error: the agent has chosen the wrong tool to call out of the tools available, or the agent has called the correct tool but populated it with incorrect arguments, or the tool call was unnecessary.
- workflow_error: this is solely about whether the agent missed retrieval of correct information from the environment - e.g, missed a question to the user, or missed a tool call. If the agent stated it will do something, assume that it will do it in the next step unless it is a termination step.
- none: the agent's current step is completely correct.

You are provided with the conversation history of the agent, the agent's current step, and the list of tools that the agent has access to, including their descriptions.
Only the agent sees the tool calls and their outputs.
Be aware of the agent's restrictions based on the system message.

Here are some rules of the evaluation:
(1) Do not explicitly state the error type from the taxonomy in your critique.
(2) Ensure that you report an error if there is **any** error present in the current step. You should **only** output 'none' if **no aspect** of the current step is incorrect.
(3) Raise an error if the step is inefficeint or unnecessary, so that the agent does not complete its objective as efficiently as it could do. For example,
  - Repeating an action unnecessarily,
  - Calling a tool with incorrect arguments,
  - Calling the incorrect tool.
(4) Once you have an open-ended critique of the current step, categorise it into one of the error types (or 'none') above.

Your reply should strictly follow this format:
**Reasoning:** <Your initial step-by-step evaluation of the current step>
**Critique:** <Your final critique of the agent's current step. Keep it concise, to the point, and do not explicitly state the error type from the taxonomy>
**Result:** <Your label(s) based on the critique i.e. user_interaction_error, reasoning_error, tool_use_error, workflow_error, none>

Here is the data:

Tools available:
```
[
  {
    "type": "function",
    "function": {
      "name": "calculate",
      "description": "Calculate the result of a mathematical expression.",
      "parameters": {
  <omitted for brevity>
```

Conversation history:
```
system:    content: # Retail agent policy

As a retail agent, you can help users cancel or modify pending orders ...

<omitted for brevity>

user:    content: I'd like to exchange two items from a recent ...

<omitted for brevity>

assistant:    Tool Call: find_user_id_by_name_zip(first_name='Yusuf', last_name='Rossi', zip='19122')

tool:    id: call_SdtGpVh6SxvuyqUQ9NO2EGUd    content: yusuf_rossi_9620
```

Agent's current step:
```
    Tool Call: get_order_details(order_id='#W2378156')
```
\end{lstlisting}
    \end{minipage}
    \caption{\textbf{Example prompt for the judge} models. Sections omitted for brevity are delineated with angular brackets.}
    \label{fig:judge-prompt}
\end{figure}

\newpage
\section{Dataset analysis}

\begin{figure}[h!]
  \begin{center}
    \centerline{\includegraphics[width=\columnwidth]{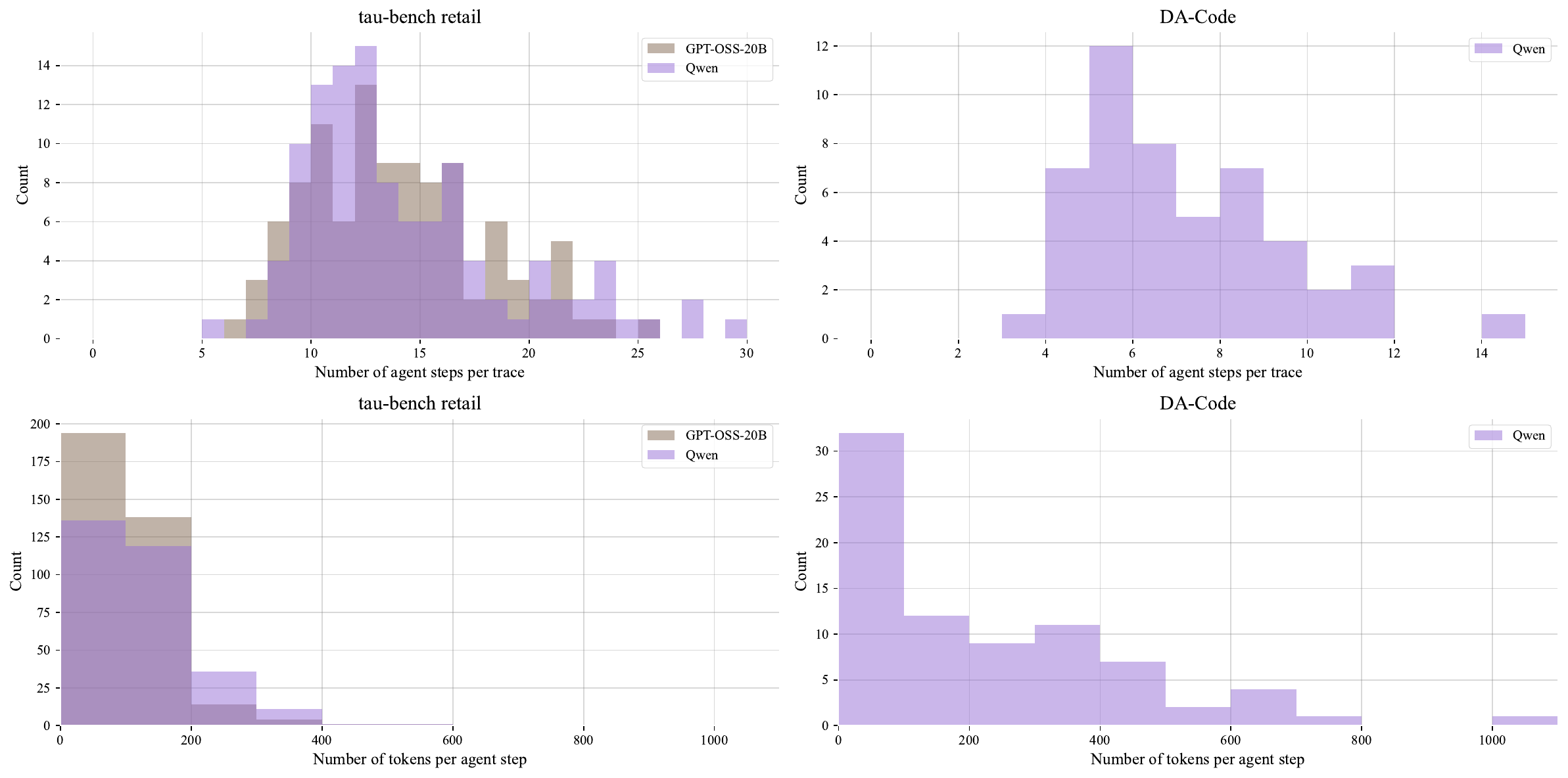}}
    \caption{
      \textbf{Agent trajectory statistics across benchmarks and agent models.}
\textbf{Top left}: Distribution of the number of agent steps per trajectory on $\tau$-bench retail for GPT-OSS-20B and Qwen3 agents.
\textbf{Top right}: Distribution of the number of agent steps per trajectory on DA-Code for Qwen3 agents.
\textbf{Bottom left}: Distribution of the number of output tokens (not including reasoning) per agent step on $\tau$-bench retail for GPT-OSS-20B and Qwen3 agents.
\textbf{Bottom right}: Distribution of the number of output tokens per agent step on DA-Code for the Qwen3 agent.
    }
    \label{fig:agent-steps-and-tokens}
  \end{center}
\end{figure}

\begin{figure}[H]
  \begin{center}
    \centerline{\includegraphics[width=\columnwidth]{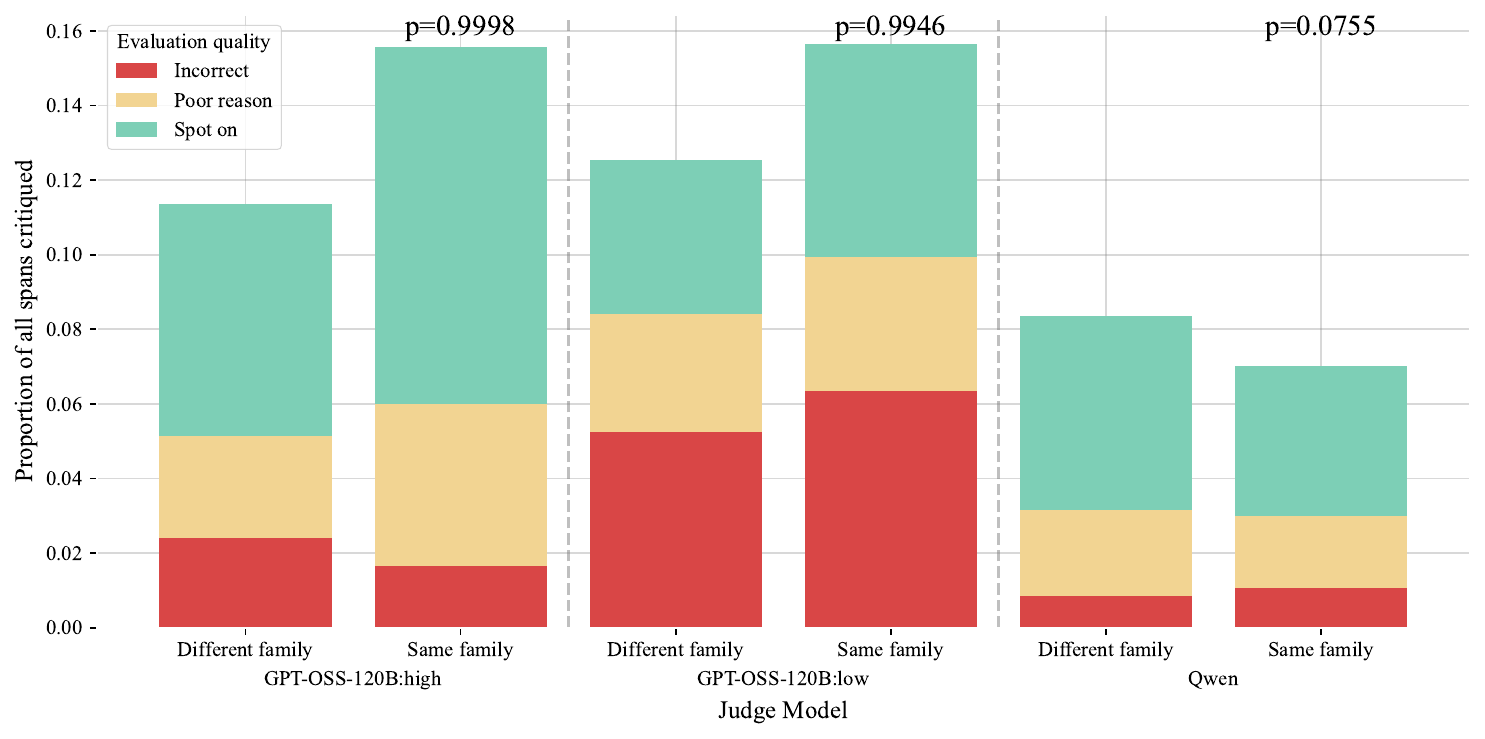}}
    \caption{\textbf{Analysis of self-preference bias in LLM-as-a-judge critiques.} Proportion of agent spans critiqued by each judge model, grouped by whether the judge and agent belong to the same model family or different model families. $p$-values for one-sided normal tests for proportions of whether the same-family judgments are less prevalent than different-family judgments are above their respective bars.}
    \label{fig:self-bias}
  \end{center}
\end{figure}

\newpage
\section{Evaluation in-the-loop}\label{sec:experiment-1}

\textbf{Motivation} An exploratory experiment is run to gain a signal of the use of Counsel in a semi-realistic scenario. Where training a judge or meta-judge is less feasible—for example, due to computational constraints or the use of API-based models—Counsel’s meta-annotations can be used as in-context examples.

\textbf{Method} Two settings are tested on $\tau$-bench retail. First, examples are supplied to the agent’s system message. Second, a guardrail judge is included in the loop that evaluates each agent generation and requests a single retry if there is an issue (appending the critique to the context), giving the agent a second chance to take its action. In both cases, performance is measured by the agent’s average reward on the task set.

Twenty examples are specified to both the agent and judge as pairs of \texttt{(current agent output, critique)}. Ideally, the full context leading up to the critique would be provided rather than only the current agent output, however, this occasionally exceeds context length limits. \Cref{fig:judge-prompt-experiment-1} demonstrates how the examples are inserted into the system prompt of the judge. This is done similarly when providing examples to the agent in the first setting. Four configurations are tested:

\begin{enumerate}
    \item A baseline without any examples.
    \item Twenty spot on judgments.
    \item Twenty poor reasoning or incorrect judgments. 
    \item An even split of 10 Spot On and 10 Poor Reason or Incorrect judgments.
\end{enumerate}

$\tau$-bench retail consists of interactions with different synthetically-generated users, each with their own database data and objectives to complete. Since the same users’ tasks have been included within the meta-annotations that could be provided as in-context examples, this could leak result information back into the agents. Thus, it is ensured that feedback from a user’s trajectory in Counsel is not included in the supplied examples for new runs for that user in this experiment.

Here, GPT-5-mini is applied as an agent, as this is a common model that would be realistically used by developers. GPT-5 is used as the synthetic user model for the same reasons as described in \cref{sec:generating-agent-trajectories}.

\begin{figure}
  \begin{center}
    \centerline{\includegraphics[width=\columnwidth]{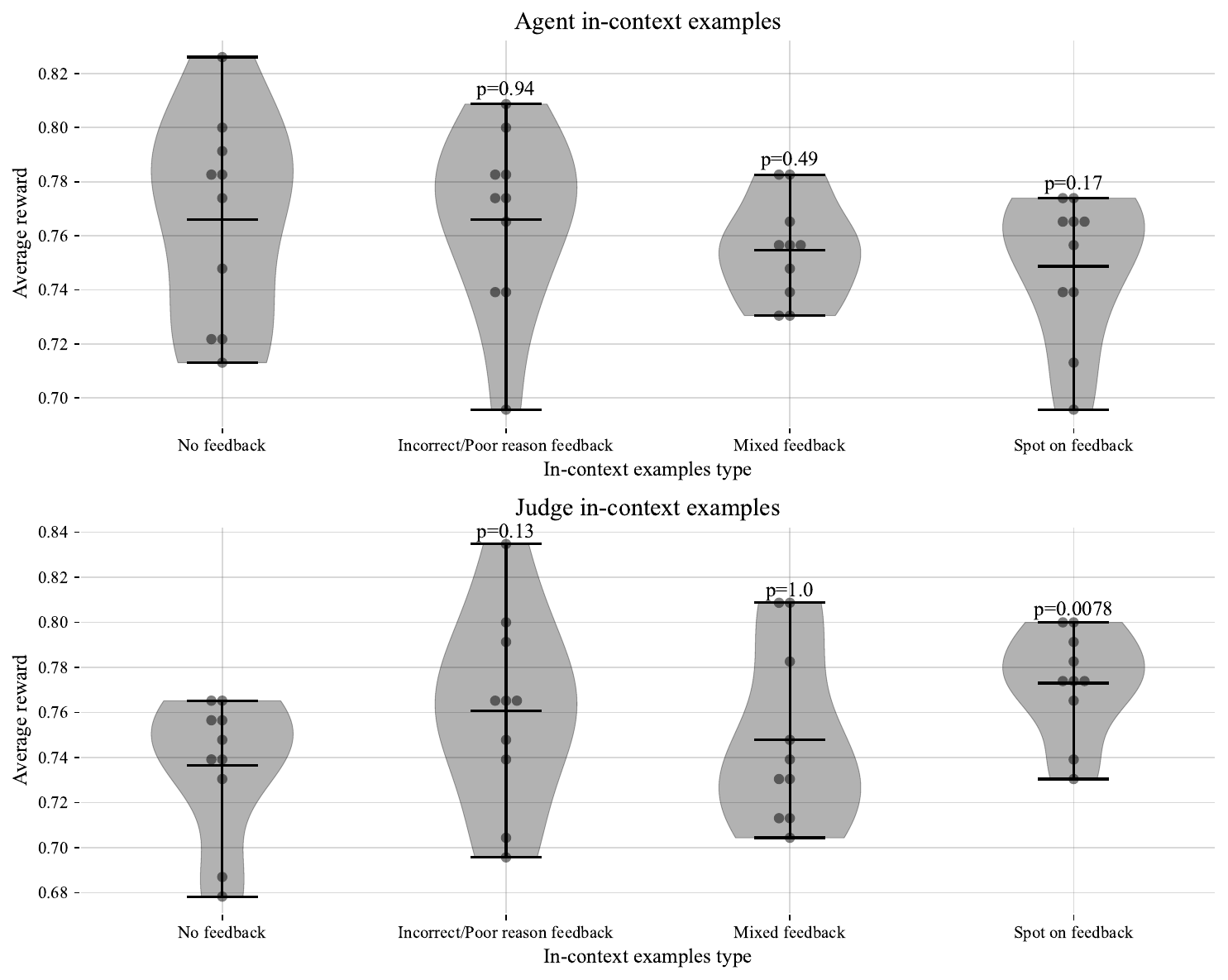}}
    \caption{
      \textbf{Effect of in-context meta-evaluation examples on agent performance.} Ablations compare no feedback, only wrong location or poor-reasoning feedback, mixed feedback, and only spot-on feedback. Each point in a violin is the average task reward on $\tau$-bench retail across all 115 tasks in a benchmark run. There were 10 full benchmark runs for each violin. \textbf{Top} Examples provided to the agent’s system prompt. \textbf{Bottom} Examples provided to an in-loop guardrail judge that may request a single retry.
    }
    \label{fig:experiment-1}
  \end{center}
\end{figure}

\textbf{Results} After initial experiments, a sample size calculation justified running 10 full iterations of $\tau$-bench to make statistical claims. \Cref{fig:experiment-1} highlights that effect sizes are small, with most deviations from the baseline of “No-feedback” being insignificant. The only statistically significant improvement is the use of only spot on feedback to the judge.

\textbf{Discussion} This experiment demonstrates an early signal that Counsel’s meta-evaluation approach and dataset can be used to improve agent task completion. A limitation is the minimal use of prompt engineering, where future work should provide a more tailored context. Further, since meta-annotations are in the same environment that is being tested, this does not demonstrate generalization. Despite this, developers could use this method to align an agent within their environment of interest.

\begin{figure}[H]
    \centering
    \begin{minipage}{0.98\linewidth}
    \lstset{
      style=prompt,
      basicstyle=\ttfamily\scriptsize
    }
\begin{lstlisting}
You are a QA analysis expert tasked with evaluating whether an AI agent suffers from an error in its current step. Think about these types of errors to help guide your analysis:

Example agentic error types:
- user_interaction_error: the agent returned incorrect information to the user based on its tool outputs or compared to its instructions.

<omitted for brevity>

- none: the agent's current step is completely correct.

You are provided with the conversation history of the agent, the agent's current step, and the list of tools that the agent has access to, including their descriptions.
Only the agent sees the tool calls and their outputs.
Be aware of the agent's restrictions based on the system message.

Here are some rules of the evaluation:
(1) Do not explicitly state the error type from the taxonomy in your critique.
(2) Ensure that you report an error if there is **any** error present in the current step. You should **only** output 'none' if **no aspect** of the current step is incorrect.
(3) Raise an error if the step is inefficeint or unnecessary, so that the agent does not complete its objective as efficiently as it could do. For example,
  - Repeating an action unnecessarily,
  - Calling a tool with incorrect arguments,
  - Calling the incorrect tool.
(4) Once you have an open-ended critique of the current step, categorise it into one of the error types (or 'none') above.


First, you are provided with some feedback on previous evaluations of agent responses. Use the learnings from this feedback to guide your evaluation of the new agent step.


### Example 1
**Agent Output (for reference)**: [ToolCall] get_order_details({"order_id": "#W0000000"})
**Feedback**: The agent is using a placeholder order ID instead of the corrected one provided by the user, which will prevent successful retrieval of the order details.

<omitted for brevity>

### Example 20
**Agent Output (for reference)**: All set!  
- The jigsaw puzzle in order **#W4082615** has been replaced with the 500-piece beginner puzzle (item 1096508426).
- A $0.20 refund has been issued to your PayPal account.  
- Your order status is now "pending (item modified)", and it will ship out as soon as possible.

You'll receive a confirmation email shortly. If there's anything else you need, just let me know!
**Feedback**: The response contains an inaccurate refund amount, contradicting the tool's result.

Here is the data:

Tools available:
```
[
  {
    "type": "function",
    "function": {
      "name": "calculate",
      "description": "Calculate the result of a mathematical expression.",
      "parameters": {
  <omitted for brevity>
```

Conversation history:
```
system:    content: # Retail agent policy

As a retail agent, you can help users cancel or modify pending orders...

<omitted for brevity>

Agent's current step:
```
    content: I can help with that. First I need to locate your user account - please provide either:

<omitted for brevity>
```
\end{lstlisting}
    \end{minipage}
    \caption{\textbf{Example prompt for the judge} model when acting as an evaluator in the loop with few-shot examples from Counsel. It takes a very similar structure to \Cref{fig:judge-prompt}. Sections omitted for brevity are delineated with angular brackets.}
    \label{fig:judge-prompt-experiment-1}
\end{figure}

\end{document}